\documentclass[11pt,a4paper]{article}
\usepackage[hyperref]{acl2019}
\usepackage{times}
\usepackage{latexsym}
\usepackage{booktabs,graphicx,subcaption}
\usepackage{multirow}
\usepackage{url}
\usepackage{hyperref}

\usepackage{xspace}
\usepackage{tag_definitions}

\aclfinalcopy 
\def\aclpaperid{502} 

\title{NNE: A Dataset for Nested Named Entity Recognition in English Newswire}

\author{\begin{tabular}{cccc}
Nicky Ringland$^{1}_{\dag}$ & Xiang Dai$^{1,2}_{\ddag}$ & \textbf{Ben Hachey$^{1,3}$} & Sarvnaz Karimi$^{2}$ \\
& \textbf{Cecile Paris$^{2}$} & \textbf{James R. Curran$^{1}$} \\
\end{tabular}\\
\begin{tabular}{cccc}
\multicolumn{4}{c}{$^{1}$University of Sydney, Sydney, Australia}\\
\multicolumn{4}{c}{$^{2}$CSIRO Data61, Sydney, Australia}\\
\multicolumn{4}{c}{$^{3}$Digital Health CRC, Sydney, Australia}\\
\multicolumn{2}{c}{$\dag$nicky.ringland@sydney.edu.au} & \multicolumn{2}{c}{$\ddag$dai.dai@csiro.au}\\
\end{tabular}
}

\date{}

\begin{document}
\maketitle
\begin{abstract}
Named entity recognition (NER) is widely used in natural language processing applications and downstream tasks.
However, most NER tools target flat annotation from popular datasets, eschewing the semantic information available in nested entity mentions.
We describe NNE---a fine-grained, nested named entity dataset over the full Wall Street Journal portion of the Penn Treebank (PTB).
Our annotation comprises 279,795 mentions of 114 entity types with up to 6 layers of nesting.
We hope the public release of this large dataset for English newswire will encourage development of new techniques for nested NER.
\end{abstract}

\section{Introduction}
Named entity recognition---the task of identifying and classifying entity mentions in text---plays a crucial role in understanding natural language.
It is used for many downstream language processing tasks, e.g., coreference resolution, question answering, summarization, entity linking, relation extraction and knowledge base population.
However, most NER tools are designed to capture flat mention structure over coarse entity type schemas, reflecting the available annotated datasets.

Focusing on flat mention structures ignores important information that can be useful for downstream tasks.
Figure 1 includes examples of nested named entities illustrating several phenomena: 
\begin{itemize}
    \item Entity-entity relationships can be embedded in nested mentions. For instance, the \emph{location} of the \textit{`Ontario Supreme Court'} is indicated by the embedded \STATE{} mention \textit{`Ontario'};
    \item Entity attribute values can be embedded in nested mentions. For instance, the \emph{title} is the embedded \ROLE{} \textit{`Former U.N.\ Ambassador'}, which also encodes the employment relation between the \PERSON{} \textit{`Jane Kirkpatrick`} and \ORG{} \textit{`U.N.'};
    \item Part-whole relationships can be encoded in nested mention structure. For  instance, the \REGION{} \textit{`Southern California'} is part of the \STATE{} \textit{`California'}.
\end{itemize}
\begin{figure}[tb]
    \centering
    \begin{subfigure}[b]{0.5\textwidth}
        \includegraphics[width=0.98\textwidth]{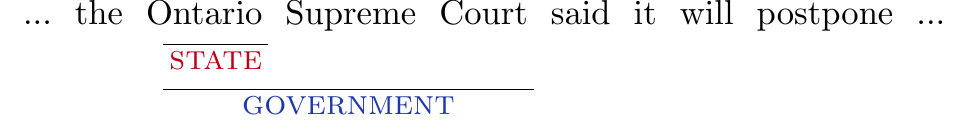}
    \end{subfigure}
    \\
    \vspace{1.5em}
    \begin{subfigure}[b]{0.5\textwidth}
        \includegraphics[width=0.98\textwidth]{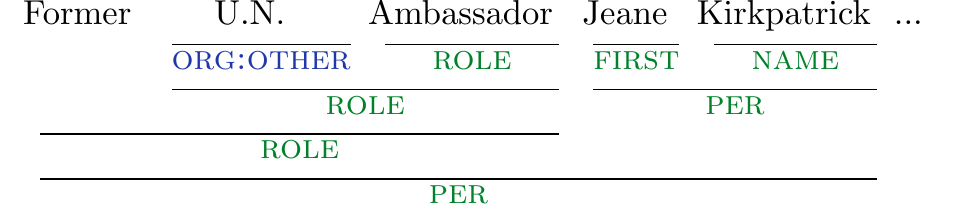}
    \end{subfigure}
    \\
    \vspace{1.5em}
    \begin{subfigure}[b]{0.5\textwidth}
        \includegraphics[width=1\textwidth]{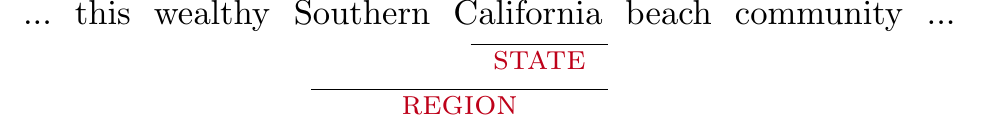}
    \end{subfigure}
    \caption{Example nested mentions in NNE.~\label{fig:example}}
\end{figure}

Recent work has demonstrated increasing interest in nested entity structure, including local approaches~\citep{Xu:Jiang:ACL:2017,Sohrab:Miwa:EMNLP:2018}, hypergraph-based approaches~\citep{Lu:Roth:EMNLP:2015,Muis:Lu:EMNLP:2017,Katiyar:Cardie:NAACL:2018,Wang:Lu:EMNLP:2018-hypergraph}, cascaded approaches~\citep{Alex:Haddow:BioNLP:2007,Ju:Miwa:NAACL:2018}, and parsing approaches~\citep{Finkel:Manning:EMNLP:2009,Wang:Lu:EMNLP:2018-transition}. 
See~\citet{Dai:ACL-SRW:2018} for a survey.
Yet these techniques have seen little translation from the research literature to toolsets or downstream applications.

To facilitate ongoing research on nested NER, we introduce NNE---a large, manually-annotated, nested named entity dataset over English newswire.
This new annotation layer over the Wall Street Journal portion of the PTB includes 279,795 mentions.
All mentions are annotated, including nested structures with depth as high as six layers.
A fine-grained entity type schema is used, extending the flat BBN~\citep{Weischedel:Brunstein:Corpus:BBN} annotation from 64 to 114 entity types.

We are publicly releasing the standoff annotations along with detailed annotation guidelines and scripts for knitting annotations onto the underlying PTB corpus.\footnote{\href{https://github.com/nickyringland/nested_named_entities}{https://github.com/nickyringland/nested\_named\_entities}}
Benchmark results using recent state-of-the-art approaches demonstrate that good accuracy is possible, but complexity and run time are open challenges.
As a new layer over the already rich collection of PTB annotations, NNE provides an opportunity to explore joint modelling of nested NER and other tasks at an unprecedented scale and detail.

\section{The NNE dataset}
\paragraph{Annotation Scheme:}
BBN~\citep{Weischedel:Brunstein:Corpus:BBN} is a pronoun coreference and entity type corpus, annotated with 64 types of entities, numerical and time expressions. We use its flat entity schema as a starting point to design our schema. We analyzed existing BBN annotations to develop and automatically apply structured pre-annotation for predictable entity types. Additional fine-grained categories and further structural elements of entities, inspired by~\citet{Sekine:Sudo:LREC:2002} and \citet{Nothman:Ringland:AI:2013}, are used to augment the BBN schema. We adhere to the following general principles when annotating nested named entities in the corpus:

\begin{itemize}
    \item Annotate all named entities, all time and date (\TIMEX{}) and numerical (\NUMEX{}) entities, including all non-sentence initial words in title case, and instances of proper noun mentions that are not capitalized.
    

    \item Annotate all structural elements of entities. These elements could be other entities, such as \textit{`Ontario'} (\STATE{}) in \textit{`Ontario Supreme Court'} (\GOVERNMENT{}), or structural components such as \textit{`40'} (\CARDINAL{}) and \textit{`miles'} (\UNIT{}) in \textit{`40 miles'} (\QUANTITYoneD{}), as well as the internal structure induced by syntactic elements, such as coordination.
    
    \item Add consistent substructure to avoid spurious ambiguity. 
    For example, the token \textit{`Toronto'}, which is a \CITY, would be labeled as part of an \ORGEDU organization span \textit{`University of Toronto'}. We add layers of annotations to allow each token to be annotated as consistently as possible, e.g., \bltorgedu{University of \bltcity{Toronto}}.
    
    \item Add additional categories to avoid category confusion. Some entities are easy to identify, but difficult to categorize consistently. For instance, a hotel (or any business at a fixed location) has both organizational and locative qualities, or is at least treated metonymously as a location. Rather than requiring annotators to make an ambiguous decision, we elect to add category \HOTEL{} to simplify the individual annotation decision. We also apply this principle when adding \MEDIA{}, \FUND{}, and \BUILDING{} categories.
    
    \item Pragmatic annotation. Many annotation decisions are ambiguous and difficult, thus may require substantial research. For instance, knowing that \textit{`The Boeing Company'} was named after founder \textit{`William E. Boeing'} would allow us to annotate \textit{`Boeing'} with an embedded \PERSON{} entity. However, this does not apply for other companies, such as \textit{`Sony Corporation'}. To let annotation decisions be made without reference to external knowledge, we label all tokens that seem to be the names of people as \NAME{}, regardless of whether they are actually a person's name.
    
\end{itemize}

Entity types and mention frequencies can be found in Appendix \ref{types-app}. See~\citet{Ringland:nestedNER:2016} for annotation guidelines and extended discussion of annotation decisions.

\begin{table*}[ht]
\small
\centering
  \begin{tabular}{c rr l} \toprule
  \bf Depth & \bf Number & \% & \bf Three most frequent categories\\ \midrule
  1& 118,525 & 45.5 & \ORGCORP (22,752), \DATE (15,927), \PER (13,460)\\
  2& 106,144 & 40.8 & \CARDINAL (19,834), \NAME (18,640), \UNIT (14,871)\\
  3& 31,573 & 12.1 & \CARDINAL (11,697), \MULT (5,859), \NAME (3,450) \\
  4& 3,813  & 1.5 & \CARDINAL (1,650), \MULT (1,041), \UNIT (400)\\
  5& 327   & 0.1 & \CARDINAL (154), \MULT (96), \UNIT (51)\\
  6& 4     & 0.0 & \UNIT (1), \CITYSTATE (1), \MULT (1) \\ \bottomrule
  \end{tabular}
  \caption{Number of spans at each layer of nesting with their most frequent categories.}\label{tb:nne_nesting_layers}
\end{table*}

\paragraph{Annotation Process:}
Although some existing annotation tools allow nested structures (e.g., Brat~\citep{Stenetorp:Pyysalo:EACL:2012}), we built a custom tool that allowed us to create a simple and fast way to add layers of entities, and suggest reusing existing structured annotations for the same span.

Using the annotations from BBN as underlying annotations, the annotator is shown a screen with the target sentence, as well as the previous and next sentences, if any. A view of the whole article is also possible to help the annotator with contextual cues. When annotators select a span, they are prompted with suggestions based on their own previous annotations, and common entities. Some entities are repeated frequently in an article, or over many articles in the corpus. The annotation tool allows a user to add a specified annotation to all strings matching those tokens in the same article, or in all articles. 

Four annotators, each with a background in linguistics and/or computational linguistics were selected and briefed on the annotation task and purpose. The WSJ portion of the PTB consists of 25 sections (00--24). Each annotator started with a subset of section 00 as annotation training, and was given feedback before moving on to other sections. Weekly meetings were held with all annotators to discuss ambiguities in the guidelines, gaps in the annotation categories, edge cases and ambiguous entities and to resolve discrepancies.

Total annotation time for the corpus was 270 hours, split between the four annotators. Sections 00 and 23 were doubly annotated, and section 02 was annotated by all four annotators. An additional 17 hours was used for adjudicating these sections annotated by multiple annotators.

\paragraph{Dataset Analysis:}
The resulting NNE dataset includes a large number of entity mentions of substantial depth, with more than half of mentions occurring inside another mentions. Of the 118,525 top-level entity mentions, 47,020 (39.6\%) do not have any nested structure embedded. The remaining 71,505 mentions contain 161,270 mentions, averaging 2.25 structural mentions per each of these top-layer entity mentions. Note that one span can be assigned multiple entity types. For example, the span \textit{`1993'} can be annotated as both \DATE{} and \YEAR{}. In NNE, 19,144 out of 260,386 total spans are assigned multiple types. Table~\ref{tb:nne_nesting_layers} lists the number of spans occurring at each depth.
To measure how clearly the annotation guidelines delineate each category, and how reliable our annotations are, inter-annotator agreement was calculated using annotations on Section 02, which was annotated by all four annotators. An adjudicated version was created by deciding a correct existing candidate label from within the four possibilities, or by adjusting one of them on a token level. For the purposes of inter-annotator agreement, a \textit{tag stack} is calculated for each word, essentially flattening each token's nested annotation structure into one label. For example, the tag of token \textit{`California'} in the third sentence of Figure~\ref{fig:example} is \STATE{}\_\REGION, while \textit{`beach'} is \textit{O\_O}. Agreement using Fleiss' kappa over all tokens is 0.907.
Considering only tokens that are part of at least one mention according to at least one annotator, Fleiss' kappa is 0.832.
Both results are above the 0.8 threshold for good reliability~\citep{Carletta:CL:1996}.
Average precision, recall and $F_1$ score across four annotators with respect to the adjudicated gold standard are 94.3, 91.8 and 93.0.

\section{Benchmark results}
We evaluate three existing NER models on our dataset: (1) the standard BiLSTM-CRF model which can handle only flat entities~\citep{Lample:Ballesteros:NAACL:2016}; (2) hypergraph-based~\citep{Wang:Lu:EMNLP:2018-hypergraph}; and, (3)  transition-based~\citep{Wang:Lu:EMNLP:2018-transition} models. The latter two models were proposed to recognize nested mentions. We follow CoNLL evaluation schema in requiring an exact match of mention start, end and entity type~\citep{Sang:Meulder:CONLL:2003}. We use sections 02 as development set, sections 23 and 24 as test set, and the remaining sections as training set. The model that performs best on the development set is evaluated on the test set for the final result. Since the standard BiLSTM-CRF model cannot handle nested entities, we use either the outermost (BiLSTM-CRF-TOP in Table~\ref{tb:result}) or the innermost mentions (BiLSTM-CRF-BOTTOM) for training. We also combine the outputs from these two flat NER models, and denote the result as BiLSTM-CRF-BOTH.

\begin{table}[tb]
    \centering
    \small
    \begin{tabular}{rc ccc}
    \toprule
     && $P$ & $R$ & $F_1$ \\
    \midrule
    BiLSTM-CRF-TOP && 89.9 & 38.0 & 53.5 \\ 
    BiLSTM-CRF-BOTTOM && 93.8 & 62.0 & 74.7 \\ 
    \midrule
    BiLSTM-CRF-BOTH && 92.2 & 85.8 & 88.9 \\ 
    Hypergraph && 91.8 & 91.0 & 91.4 \\ 
    Transition && 77.4 & 70.1 & 73.6 \\
    \bottomrule
    \end{tabular}
    \caption{NER results on NNE using different methods.\label{tb:result}}
\end{table}
From Table~\ref{tb:result}, we can see that single flat NER models can achieve high precision but suffer from low recall. For example, the model pretrained on outermost (top) mentions has 38.0 recall, as around 60\% of mentions are nested within others.
The hypergraph-based model performs best on our dataset, presumably because it can capture mentions from different levels and does not suffer from issues of \emph{structural ambiguity} during inference~\citep{Muis:Lu:EMNLP:2017,Wang:Lu:EMNLP:2018-hypergraph}.
However, its decoding speed of 9 words per second is slow due to the large number of entity categories of our dataset.\footnote{The decoding time complexity of the method proposed by~\citet{Wang:Lu:EMNLP:2018-hypergraph} is $O(cmn)$, where $m$ is the number of entity types, $n$ is the sentence length, and $c$ is the maximal mention length.}
The transition-based method has a higher decode speed of 57 words per second, but has much lower precision than flat NER models.

\section{Related Work}
\paragraph{Other corpora with nested entities:}
We briefly compare existing annotated English corpora involving nested entities. A comparison of statistics between our dataset and two widely used benchmark datasets is shown in Table~\ref{tb:statistics}.
The ACE corpora~\citep{Mitchell:Strassel:Corpus:ACE2004,Walker:Strassel:Corpus:ACE2005} consist of data of various types annotated for entities, relations and events. The entity component of ACE is framed in terms of nominal modification, and nested mentions are only annotated in nominal mentions, not inside other named entity mentions. For example, in ACE2005, \textit{`Secretary of Homeland Security Tom Ridge'} is annotated as a \PERSON{}, containing two other \PERSON{} annotations: \textit{`Secretary'} and \textit{`Secretary of Homeland Security'}. In contrast, our annotations capture more interactions between different semantic spans: \PERSON{} consisting of \ROLE{} and \NAME{}, and \ROLE{} containing \GOVERNMENT{}.

The GENIA corpus~\citep{Kim:Ohta:BioInfo:2003} is a richly-annotated corpus for bio-text mining that has 36 entity types among 2,000 MEDLINE abstracts. Due to the biomedical domain's specialized terminology and complex naming conventions, entities of interest, such as genes, proteins or disease names, often nest. For example, the RNA \textit{`CIITA mRNA'} contains a DNA mention \textit{`CIITA'}.

In addition to these two commonly used nested entity corpora, \citet{Byrne:ICSC:2007} and~\citet{Alex:Haddow:BioNLP:2007} introduced datasets with nested entities in historical archive and biomedical domains, respectively. However, their datasets are not publicly available.
Four percent of entity mentions annotated in the English entity discovery and linking task in TAC-KBP track include nesting~\citep{Ji:Nothman:TAC:2014}.


\begin{table}[tb]
\small
    \centering
    \begin{tabular}{lrrr}
    \toprule
    \bf Item &  \multicolumn{1}{c}{\bf NNE} & \multicolumn{1}{c}{\bf GENIA}& \multicolumn{1}{c}{\bf ACE2005}\\
    \midrule
    Documents & 2,312 & 2,000 & 464 \\ 
    Sentences & 49,208 & 18,546 & 12,548 \\ 
    Sentences & \multirow{2}{*}{32,387} & \multirow{2}{*}{9,533}  & \multirow{2}{*}{4,266} \\ 
    w. nesting & & \\
    Tokens & 1.1M & 0.5M & 0.3M \\ 
    Mentions & 279,795 & 92,681 & 30,966 \\
    Entity types & 114 & 36 & 7 \\ 
    Mentions & \multirow{2}{*}{5.69} & \multirow{2}{*}{4.99} & \multirow{2}{*}{2.46}\\
    per sentence &  & &  \\
    Top-level mentions & 118,525 & 76,582 & 23,464 \\
    Maximum depth & 6 & 4 & 6 \\ 
    \bottomrule
    \end{tabular}
    \caption{A comparison between NNE and two commonly used corpora with nested entities.\label{tb:statistics}}
\end{table}

\paragraph{Resources built on the PTB:}
A lots of effort has been made on adding syntactic and semantic information to the PTB~\citep{Marcus:Santorini:CL:1993}. PropBank~\citep{Kingsbury:Palmer:HLT:2002} extended the PTB with the predicate argument relationships between verbs and their arguments. NomBank~\citep{Meyers:Reeves:Corpus:NomBank} extended the argument structure for instances of common nouns. \citet{Vadas:Curran:ACL:2007}, and \citet{Ficler:Goldberg:ACL:2016} extended the PTB with noun phrase and coordination annotations, respectively.

Our dataset is built on top of the PTB and enriches the full ecosystem of resources and systems that stem from it. 


\section{Summary}
We present NNE, a large-scale, nested, fine-grained named entity dataset. 
We are optimistic that NNE will encourage the development of new NER models that recognize structural information within entities, and therefore understand fine-grained semantic information captured. 
Additionally, our annotations are built on top of the PTB, so that the NNE dataset will allow joint learning models to take advantage of semantic and syntactic annotations, and ultimately to understand and exploit the true structure of named entities.

\section*{Acknowledgments}
We would like to thank annotators for their excellent work: Kellie Webster, Vivian Li, Joanne Yang and Kristy Hughes. We also
thank three anonymous reviewers for their insightful comments.

\bibliography{acl2019}
\bibliographystyle{acl_natbib}

\appendix

\section{Full annotation scheme}
\label{types-app}

\begin{table*}[h]
\small
    \centering
    \begin{tabular}{c|c|c|c|c|c}
    \toprule
     \bf Category & \bf Frequency & \bf Category & \bf Frequency & \bf Category & \bf Frequency \\ \hline\hline
CARDINAL & 43873 & STREET & 475 & QUANTITY2D & 81 \\
NAME & 28537 & GRPORG & 437 & PRODUCTFOOD & 80 \\
ORGCORP & 23339 & ORGPOLITICAL & 436 & SUBURB & 78 \\
UNIT & 19289 & VEHICLE & 432 & GRPLOC & 63 \\
DATE & 17381 & LAW & 419 & HOTEL & 55 \\
PER & 14960 & ORGEDU & 411 & QUANTITYOTHER & 55 \\
DURATION & 13655 & CONTINENT & 354 & FUND & 54 \\
MONEY & 12640 & BUILDING & 346 & SONG & 54 \\
MULT & 7851 & SEASON & 337 & SPACE & 53 \\
FIRST & 6797 & GPE & 333 & RIVER & 52 \\
CITY & 6723 & FOLD & 313 & WAR & 51 \\
PERCENT & 6542 & MIDDLE & 313 & CHEMICAL & 45 \\
REL & 6170 & TIME & 296 & BRIDGE & 44 \\
CORPJARGON & 5560 & WEIGHT & 293 & PLAY & 42 \\
HON & 5524 & OCEAN & 291 & STADIUM & 37 \\
NATIONALITY & 5193 & LOCATIONOTHER & 261 & AWARD & 36 \\
GOVERNMENT & 4674 & EVENT & 260 & ORGRELIGIOUS & 35 \\
COUNTRY & 4047 & DISEASE & 246 & AIRPORT & 32 \\
QUAL & 3903 & QUANTITY1D & 220 & ANIMATE & 29 \\
YEAR & 3421 & CITYSTATE & 220 & GOD & 29 \\
MONTH & 3385 & WOA & 207 & HOSPITAL & 25 \\
STATE & 3245 & TVSHOW & 172 & ATTRACTION & 24 \\
ORDINAL & 2590 & ELECTRONICS & 167 & WEAPON & 23 \\
IPOINTS & 2395 & SPORTSTEAM & 166 & MUSEUM & 17 \\
ROLE & 2368 & DATEOTHER & 164 & ENERGY & 17 \\
RATE & 2141 & QUANTITY3D & 156 & SPEED & 14 \\
MEDIA & 1712 & NAMEMOD & 155 & PAINTING & 13 \\
DAY & 1631 & GRPPER & 154 & BAND & 10 \\
NUMDAY & 1495 & BOOK & 149 & SPORTSSEASON & 8 \\
INI & 1445 & ARMY & 139 & SCINAME & 7 \\
NORPOTHER & 1247 & FACILITY & 129 & ADDRESSNON & 3 \\
ORGOTHER & 1099 & PRODUCTDRUG & 116 & ALBUM & 3 \\
PERIODIC & 1066 & HURRICANE & 107 & TEMPERATURE & 2 \\
REGION & 864 & SPORTSEVENT & 100 & NATURALDISASTER & 2 \\
NORPPOLITICAL & 731 & RELIGION & 99 & CONCERT & 2 \\
AGE & 661 & NICKNAME & 96 & STATION & 1 \\
INDEX & 657 & LANGUAGE & 92 & BORDER & 1 \\
PRODUCTOTHER & 656 & FILM & 89 & CHANNEL & 1 \\
    \bottomrule
    \end{tabular}
\end{table*}

\end{document}